\documentclass[final]{cvpr}

\usepackage{times}
\usepackage{epsfig}
\usepackage{graphicx}
\usepackage{amsmath}
\usepackage{amssymb}

\usepackage[pagebackref,breaklinks,colorlinks]{hyperref}

\usepackage{booktabs}
\usepackage{multicol}
\usepackage{multirow}
\usepackage{color}
\usepackage{caption}
\usepackage{subcaption}
\usepackage{hhline}
\usepackage{pifont}
\usepackage{threeparttable}
\usepackage{makecell}

\def\cvprPaperID{****} 
\def\confName{CVPR}
\def\confYear{2022}
\graphicspath{{./figs/}{./figs/result/}}

\begin{document}

\title{PyramidTNT: Improved Transformer-in-Transformer Baselines\\with Pyramid Architecture}

\author{Kai Han, Jianyuan Guo, Yehui Tang, Yunhe Wang\\
	Huawei Noah's Ark Lab\\
	{\tt\small \{kai.han,jianyuan.guo,tangyehui,yunhe.wang\}@huawei.com}
}
\maketitle

\begin{abstract}
	Transformer networks have achieved great progress for computer vision tasks. Transformer-in-Transformer (TNT) architecture utilizes inner transformer and outer transformer to extract both local and global representations. In this work, we present new TNT baselines by introducing two advanced designs: 1) pyramid architecture, and 2) convolutional stem. The new ``PyramidTNT'' significantly improves the original TNT by establishing hierarchical representations. PyramidTNT achieves better performances than the previous state-of-the-art vision transformers such as Swin Transformer. We hope this new baseline will be helpful to the further research and application of vision transformer. Code will be available at \url{https://github.com/huawei-noah/CV-Backbones/tree/master/tnt_pytorch}.
\end{abstract}

\section{Introduction}\label{sec:introduction}
Vision transformer is providing a new type of neural network for computer vision. Starting from ViT~\cite{vit}, a series of works have been proposed to improve the architecture of vision transformer~\cite{tnt,yuan2021tokens,wang2021pyramid,levit,liu2021swin,chu2021twins,augvit}. PVT~\cite{wang2021pyramid} introduces pyramid network architecture for vision transformer. T2T-ViT-14~\cite{yuan2021tokens} recursively aggregates neighboring tokens into one token for extracting local structure and reducing the number of tokens. TNT~\cite{tnt} utilizes inner transformer and outer transformer to model word-level and sentence-level visual representations.  Swin Transformer~\cite{liu2021swin} proposes a hierarchical transformer whose representation is computed with Shifted windows. With the recent progress, the performance of vision transformer shows superiority over convolutional neural network (CNN)~\cite{survey}.

This work establishes improved vision transformer baselines based on the TNT~\cite{tnt} framework. Inspired by the recent works~\cite{wang2021pyramid,xiao2021early}, we introduce two main architecture modifications: 1) pyramid architecture with gradual decreased resolution to extract multi-scale representations, and 2) convolutional stem for improving the patchify stem and stable training. We also include several other tricks~\cite{shaw2018self,pvtv2} to further improve the efficiency. The new transformer is named as PyramidTNT. The experiments on image classification and object detection demonstrate the superiority of PyramidTNT. Specifically, PyramidTNT-S yields 82.0\% ImageNet classification top-1 accuracy with only 3.3B FLOPs, which is significantly better than the original TNT-S~\cite{tnt} and Swin-T~\cite{liu2021swin}. For COCO detection, PyramidTNT-S achieves 42.0 mAP with fewer computational cost than othere transformer and MLP detection models. We hope this new baseline will be helpful to the further research and application of vision transformer.

\begin{figure}[tp]
	\centering	
	\setlength{\tabcolsep}{3pt}{
		\renewcommand{\arraystretch}{1.0}
		\begin{tabular}{cc}
			\makecell*[c]{\includegraphics[width=0.48\linewidth]{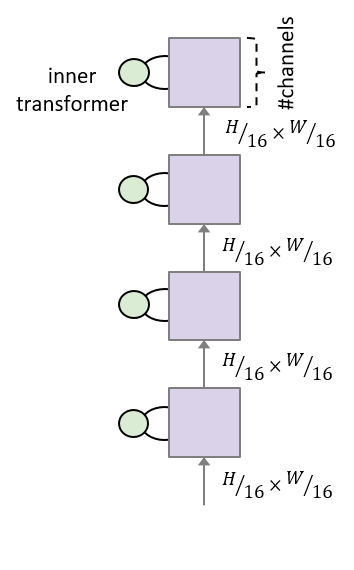}}  & \makecell*[c]{\includegraphics[width=0.48\linewidth]{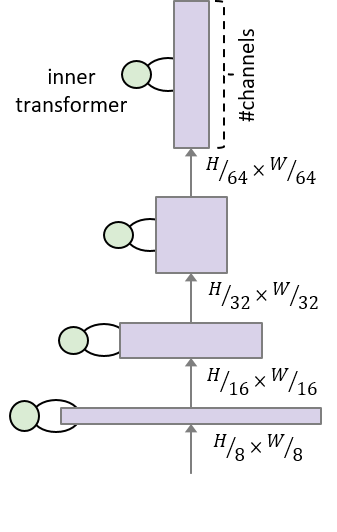}} 
			\\
			\small (a) TNT & \small (b) PyramidTNT
		\end{tabular}
	}
	\vspace{-0.5em}
	\caption{Comparison of TNT and PyramidTNT architectures.}
	\label{Fig:ptnt}
	\vspace{-1.0em}
\end{figure}

\begin{table*}[htp]
	\centering
	\scriptsize
	\caption{Network architectures of PyramidTNT. Three instantiations with different complexity including tiny (Ti), small (S), middle (M) and base (B) versions are presented. The expansion ratio of MLP module is set as 4 by default. $H_o$ and $H_i$ denote the number of heads in outer transformer and inner transformer. $R$ is the reduction ratio of the LSRA.}
	\label{table-arch}
	\vspace{-1em}
	\renewcommand{\arraystretch}{1.15}
	\setlength{\tabcolsep}{1.75pt}
	\begin{tabular}{c|c|c|c|c|c|c|c|c|c}
		\Xhline{1.2pt}
		Stage & Output size & \multicolumn{2}{c|}{PyramidTNT-Ti} & \multicolumn{2}{c|}{PyramidTNT-S} & \multicolumn{2}{c|}{PyramidTNT-M} & \multicolumn{2}{c}{PyramidTNT-B}
		\\
		\hline
		Stem & $\frac{H}{8}\times\frac{W}{8}$ & \multicolumn{2}{c|}{Conv$\times5$} & \multicolumn{2}{c|}{Conv$\times5$} & \multicolumn{2}{c|}{Conv$\times5$} & \multicolumn{2}{c}{Conv$\times5$}
		\\
		\hline
		\multirow{4}{*}{Stage 1} & \multirow{4}{*}{$\frac{H}{8}\times\frac{W}{8}$} & Outer & Inner & Outer & Inner & Outer & Inner & Outer & Inner
		\\
		& & $\begin{bmatrix}D=80\\H_{o}=2\\R=4\end{bmatrix}$$\times$2 
		& $\begin{bmatrix}C=5\\H_{i}=1\\R=1\end{bmatrix} $$\times$1
		& $\begin{bmatrix}D=128\\H_{o}=4\\R=4\end{bmatrix} $$\times$2 
		& $\begin{bmatrix}C=8\\H_{i}=2\\R=1\end{bmatrix} $$\times$1
		& $\begin{bmatrix}D=192\\H_{o}=4\\R=4\end{bmatrix} $$\times$2
		& $\begin{bmatrix}C=12\\H_{i}=2\\R=1\end{bmatrix}  $$\times$1
		& $\begin{bmatrix}D=256\\H_{o}=4\\R=4\end{bmatrix} $$\times$2
		& $\begin{bmatrix}C=16\\H_{i}=2\\R=1\end{bmatrix} $$\times$1
		\\
		\hline
		Downsample & $\frac{H}{16}\times\frac{W}{16}$ & \multicolumn{2}{c|}{Patch Merging} & \multicolumn{2}{c|}{Patch Merging} & \multicolumn{2}{c|}{Patch Merging} & \multicolumn{2}{c}{Patch Merging}
		\\
		\hline
		\multirow{4}{*}{Stage 2} & \multirow{4}{*}{$\frac{H}{16}\times\frac{W}{16}$} & Outer & Inner & Outer & Inner & Outer & Inner & Outer & Inner
		\\
		& & $\begin{bmatrix}D=160\\H_{o}=4\\R=2\end{bmatrix}$$ \times$6
		& $\begin{bmatrix}C=10\\H_{i}=2\\R=1\end{bmatrix} $$\times$1
		& $\begin{bmatrix}D=256\\H_{o}=8\\R=2\end{bmatrix}$$ \times$8
		& $\begin{bmatrix}C=16\\H_{i}=4\\R=1\end{bmatrix} $$\times$2
		& $\begin{bmatrix}D=384\\H_{o}=8\\R=2\end{bmatrix} $$\times$8
		& $\begin{bmatrix}C=24\\H_{i}=4\\R=1\end{bmatrix} $$ \times$2
		& $\begin{bmatrix}D=512\\H_{o}=8\\R=2\end{bmatrix} $$\times$10
		& $\begin{bmatrix}C=32\\H_{i}=4\\R=1\end{bmatrix} $$\times$2
		\\\hline
		Downsample & $\frac{H}{32}\times\frac{W}{32}$ & \multicolumn{2}{c|}{Patch Merging} & \multicolumn{2}{c|}{Patch Merging} & \multicolumn{2}{c|}{Patch Merging} & \multicolumn{2}{c}{Patch Merging}
		\\
		\hline
		\multirow{4}{*}{Stage 3} & \multirow{4}{*}{$\frac{H}{32}\times\frac{W}{32}$} & Outer & Inner & Outer & Inner & Outer & Inner & Outer & Inner
		\\
		& & $\begin{bmatrix}D=320\\H_{o}=8\\R=1\end{bmatrix} $$\times$3
		& $\begin{bmatrix}C=20\\H_{i}=4\\R=1\end{bmatrix} $$\times$1
		& $\begin{bmatrix}D=512\\H_{o}=16\\R=1\end{bmatrix}$$ \times$4
		& $\begin{bmatrix}C=32\\H_{i}=8\\R=1\end{bmatrix}$$ \times$1
		& $\begin{bmatrix}D=768\\H_{o}=16\\R=1\end{bmatrix} $$\times$6
		& $\begin{bmatrix}C=48\\H_{i}=8\\R=1\end{bmatrix} $$ \times$1
		& $\begin{bmatrix}D=1024\\H_{o}=16\\R=1\end{bmatrix} $$\times$6
		& $\begin{bmatrix}C=64\\H_{i}=8\\R=1\end{bmatrix} $$\times$1
		\\
		\hline
		Downsample & $\frac{H}{64}\times\frac{W}{64}$ & \multicolumn{2}{c|}{Patch Merging} & \multicolumn{2}{c|}{Patch Merging} & \multicolumn{2}{c|}{Patch Merging} & \multicolumn{2}{c}{Patch Merging}
		\\
		\hline
		\multirow{4}{*}{Stage 4} & \multirow{4}{*}{$\frac{H}{64}\times\frac{W}{64}$} & Outer & Inner & Outer & Inner & Outer & Inner & Outer & Inner
		\\
		& & $\begin{bmatrix}D=320\\H_{o}=8\\R=1\end{bmatrix}$$ \times$2
		& $\begin{bmatrix}C=20\\H_{i}=4\\R=1\end{bmatrix}$$ \times$1
		& $\begin{bmatrix}D=512\\H_{o}=16\\R=1\end{bmatrix}$$ \times$2
		& $\begin{bmatrix}C=32\\H_{i}=8\\R=1\end{bmatrix}$$ \times$1
		& $\begin{bmatrix}D=768\\H_{o}=16\\R=1\end{bmatrix}$$ \times$2
		& $\begin{bmatrix}C=48\\H_{i}=8\\R=1\end{bmatrix} $$ \times$1
		& $\begin{bmatrix}D=1024\\H_{o}=16\\R=1\end{bmatrix}$$ \times$2
		& $\begin{bmatrix}C=64\\H_{i}=8\\R=1\end{bmatrix} $$\times$1
		\\
		\hline
		Head & $1\times1$ & \multicolumn{2}{c|}{Pooling \& FC} & \multicolumn{2}{c|}{Pooling \& FC} & \multicolumn{2}{c|}{Pooling \& FC} & \multicolumn{2}{c}{Pooling \& FC}\\
		\hline
		\multicolumn{2}{c|}{Input resolution} & \multicolumn{2}{c|}{192$\times$192} & \multicolumn{2}{c|}{256$\times$256} & \multicolumn{2}{c|}{256$\times$256} & \multicolumn{2}{c}{256$\times$256}
		\\
		\hline
		\multicolumn{2}{c|}{Parameters (M)} & \multicolumn{2}{c|}{10.6} & \multicolumn{2}{c|}{32.0} & \multicolumn{2}{c|}{85.0} & \multicolumn{2}{c}{157.0}
		\\
		\hline
		\multicolumn{2}{c|}{FLOPs (B)} & \multicolumn{2}{c|}{0.6} & \multicolumn{2}{c|}{3.3} & \multicolumn{2}{c|}{8.2} & \multicolumn{2}{c}{16.0}
		\\
		\Xhline{1.2pt}
	\end{tabular}
	\vspace{-1em}
\end{table*}

\section{Related Work}\label{sec:related}

\paragraph{Transformer Backbone.} Dosovitskiy~\etal~\cite{vit} firstly introduce the pure transformer architecture~\cite{vaswani2017attention} to the vision tasks, which splits the input image into multiple patches and takes each patch as a `word' in natural language. In \cite{vaswani2017attention}, extremely large training datasets (\eg, JFT-300M and ImageNet-21k) are usually required for high performance. Touvron~\etal~\cite{DeiT} improve the training recipe and train vision transformers from scratch on ImageNet. Wang~\etal~\cite{wang2021pyramid} introduce a hierarchical architecture, which reduce the sequence length of transformer as the network deepens, which can extract the high-level semantic information and reduce the computational cost.  Liu~\etal~\cite{liu2021swin} restrict the self-attention operation in non-overlapping local windows and realize the cross-window connection by shifting these windows. Yuan~\etal~\cite{yuan2021tokens} propose a layer-wise tokens-to-token transformation to replace the simple tokenization of input images. Wu~\etal~\cite{wu2021cvt} introduce convolutional projections into vision transformers to bring the  desirable properties of CNNs.
To capture both global and local information in an image, Han~\etal~\cite{tnt} present a nested architecture by further dividing each patch into smaller ones, which enhances the representation ability significantly. Notice that this nested design is a general  methodology, which can  be also combined with the hierarchical architectures (\eg, \cite{wang2021pyramid, liu2021swin}) for  further improving performance. 

\paragraph{MLP Backbone.}
Tolstikhin~\etal~\cite{tolstikhin2021mlp} construct a MLP-Mixer model by only stacking multi-layer perceptrons (MLPs),  showing that neither  convolutions and attention are necessary for good performance. Channel-mixing and token-mixing MLPs are two core blocks, which extract features of each token (patch) and aggregate information from different tokens, respectively. Recently, various variants are developed to achieve a better trade-off between accuracy and computational cost.  For example, shift operation is introduced in $S^2$-MLP~\cite{yu2021s} and AS-MLP~\cite{asmlp} to exchange information across different tokens. Hire-MLP~\cite{hire} present a hierarchical rearrangement operation, where the inner-region rearrangement and cross-region rearrangement capture local information and global context, respectively. Wave-MLP~\cite{wavemlp} takes each tokens as a wave and model the relationship between different tokens by considering their amplitude and phase information simultaneously.

\section{Method}\label{sec:method}

\paragraph{Convolutional Stem.}
Given a input image $X\in\mathbb{R}^{H\times W\time 3}$, vanilla TNT model first split the image into a number of patches and further view each patch as a sequence of sub-patches. A linear layer is applied to project the sub-patch into a visual word vector (a.k.a., token). These visual words are concatenated and transformed into a visual sentence vector. Xiao \emph{et al.}~\cite{xiao2021early} find that using several convolutions as stem in ViT increases optimization stability and also improves the performance. Based on the observation, we construct a convolutional stem for PyramidTNT. A stack of 3$\times$3 convolutions is utilized to produce visual words $Y\in\mathbb{R}^{\frac{H}{2}\times \frac{W}{2}\times C}$ where $C$ is the visual word dimension. Similarly, we can obtain visual sentences $Z\in\mathbb{R}^{\frac{H}{8}\times \frac{W}{8}\times D}$ where $D$ is the visual sentence dimension. The word-level and sentence-level position encodings are added on visual words and sentences respectively, as in the original TNT~\cite{tnt}.

\paragraph{Pyramid Architecture.}
The original TNT network maintains the same number of tokens in every block, following ViT~\cite{vit}. The numbers of visual words and visual sentences are kept unchanged from bottom to top.
Inspired by PVT~\cite{wang2021pyramid}, we construct four stages with different number of tokens for TNT, as shown in Figure~\ref{Fig:ptnt} (b). For the four stages, the spatial shape of visual words are set as $\frac{H}{2}\times \frac{W}{2}$, $\frac{H}{4}\times \frac{W}{4}$, $\frac{H}{8}\times \frac{W}{8}$, and $\frac{H}{16}\times \frac{W}{16}$. The spatial shape of visual sentences are set as $\frac{H}{8}\times \frac{W}{8}$, $\frac{H}{16}\times \frac{W}{16}$, $\frac{H}{32}\times \frac{W}{32}$, and $\frac{H}{64}\times \frac{W}{64}$. The downsample operation is implemented using a convolution with stride 2. Each stage is composed by several TNT blocks, and the TNT block operates on word-level and sentence-level features as described in~\cite{tnt}. Finally, the output visual sentences are fused into a vector as the image representation using the global average pooling operation.

\paragraph{Other Tricks.}
In apart from the network architecture modification, several advanced tricks for vision transformer are also adopted. 
Relative position encoding~\cite{shaw2018self} is added on self-attention module to better represent relative position between tokens.
Linear spatial reduction attention (LSRA)~\cite{pvtv2} is utilized in the first two stages to reduce the computation cost of self-attention for long sequence.

\section{Experiment}\label{sec:experiment}

\subsection{Image Classification}

\begin{table}[htp]
	\centering
	\small
	\caption{Training hyperparameters for ImageNet-1K.}
	\label{table-hyper}
	\vspace{-1em}
	\setlength{\tabcolsep}{8pt}
	\begin{tabular}{l|cccc}
		\Xhline{1.2pt}
		PyramidTNT & Ti & S & M & B \\ 
		\hline
		Epochs & \multicolumn{4}{c}{300} \\
		Batch size & \multicolumn{4}{c}{1024} \\
		Optimizer & \multicolumn{4}{c}{AdamW~\cite{adamw}} \\
		Start learning rate (LR) & \multicolumn{4}{c}{1e-3} \\
		LR decay & \multicolumn{4}{c}{Cosine} \\
		Warmup epochs & \multicolumn{4}{c}{20} \\
		Weight decay & \multicolumn{4}{c}{0.05} \\
		Label smoothing~\cite{label-smooth} & \multicolumn{4}{c}{0.1} \\
		Drop path~\cite{larsson2016fractalnet} & 0.1 & 0.1 & 0.15 & 0.3 \\
		Repeated augment~\cite{hoffer2020augment} & \multicolumn{4}{c}{$\checkmark$} \\
		RandAugment~\cite{randaugment} & \multicolumn{4}{c}{$\checkmark$} \\
		Mixup prob.~\cite{mixup} & \multicolumn{4}{c}{0.8} \\
		Cutmix prob.~\cite{cutmix} & \multicolumn{4}{c}{1.0} \\
		Erasing prob.~\cite{erasing} & \multicolumn{4}{c}{0.25} \\
		Exponential moving average & \multicolumn{4}{c}{0.99996} \\
		\Xhline{1.2pt}
	\end{tabular}
	\vspace{-1em}
\end{table}

\paragraph{Settings.}
We conduct image classification experiments on the large-scale ImageNet-1K dataset~\cite{imagenet}. ImageNet-1K consists of about 1.28M training images and 50K validation images belonging to 1,000 classes. We utilize the same training strategy as in DeiT~\cite{DeiT} and TNT~\cite{tnt}, as described in Table~\ref{table-hyper}. All PyramidTNT models are implemented using PyTorch and trained on 8 NVIDIA V100 GPUs.

\begin{table}[]
	\centering
	\small
	\caption{ImageNet-1K classification results of representative CNN, MLP and transformer models. Following~\cite{DeiT,liu2021swin}, the throughput is measured on an NVIDIA V100 GPU and PyTorch.}
	\label{table-imagenet}
	\vspace{-1em}
	\setlength{\tabcolsep}{3pt}
	\begin{tabular}{l|c|c|c|c}
		\Xhline{1.2pt}
		\multirow{2}{*}{Model} & Params & FLOPs & Throughput & Top-1 \\ 
		& (M) & (B) & (image/s) & (\%) \\
		\hline
		\multicolumn{5}{c}{\textbf{CNN}} \\
		\hline
		ResNet-50~\cite{resnet,yuan2021tokens} & 25.6 & 4.1 & 1226 & 79.1 \\
		ResNet-101~\cite{resnet,yuan2021tokens} & 44.7 & 7.9  & 753 & 79.9  \\
		ResNet-152~\cite{resnet,yuan2021tokens} & 60.2 & 11.5  & 526 & 80.8  \\  
		\hline
		EfficientNet-B0~\cite{efficientnet} & 5.3 & 0.39  & 2694 & 77.1 \\
		EfficientNet-B1~\cite{efficientnet} & 7.8 & 0.7  & 1662 & 79.1 \\
		EfficientNet-B2~\cite{efficientnet} & 9.2 & 1.0  & 1255 & 80.1 \\
		EfficientNet-B3~\cite{efficientnet} & 12 & 1.8  & 732 & 81.6 \\
		EfficientNet-B4~\cite{efficientnet} & 19 & 4.2  & 349 & 82.9 \\
		\hline
		S-GhostNet-B1~\cite{liu2021greedy,ghostnet} & 16.2 & 0.67  & - & 80.9 \\
		S-GhostNet-B4~\cite{liu2021greedy,ghostnet} & 32.9 & 4.4  & - & 84.3 \\
		
		\hline
		\multicolumn{5}{c}{\textbf{MLP}} \\
		\hline
		AS-MLP-T~\cite{asmlp}  & 28 & 4.4  & 862 & 81.3 \\
		AS-MLP-S~\cite{asmlp}  & 50 & 8.5  & 473 & 83.1 \\
		AS-MLP-B~\cite{asmlp}  & 88 & 15.2  & 308 & 83.3 \\
		\hline
		CycleMLP-B2~\cite{cyclemlp}  & 27 & 3.9  & 635 & 81.6 \\
		CycleMLP-B3~\cite{cyclemlp}  & 38 & 6.9  & 371 & 82.4 \\
		CycleMLP-B4~\cite{cyclemlp}  & 52 & 10.1  & 259 & 83.0 \\
		\hline
		Hire-MLP-Small~\cite{hire}  & 33 & 4.2  & 807 & 82.1 \\
		Hire-MLP-Base~\cite{hire}  & 58 & 8.1 & 441 & 83.2 \\
		Hire-MLP-Large~\cite{hire}  & 96 & 13.4 & 290 & 83.8 \\
		\hline
		Wave-MLP-T~\cite{wavemlp}  & 17 & 2.4  & 1208 & 80.6 \\
		Wave-MLP-S~\cite{wavemlp}  & 30 & 4.5 & 720 & 82.6 \\
		Wave-MLP-M~\cite{wavemlp}  & 44 & 7.9 & 413 & 83.4 \\
		
		\hline
		\multicolumn{5}{c}{\textbf{Transformer}} \\
		\hline
		DeiT-Ti~\cite{vit,DeiT}  & 5 & 1.3   & 2536 & 72.2 \\
		DeiT-S~\cite{vit,DeiT}   & 22 & 4.6  & 940 & 79.8 \\
		DeiT-B~\cite{vit,DeiT}  & 86 & 17.6   & 292 & 81.8 \\  
		\hline
		T2T-ViT-14~\cite{yuan2021tokens}  & 21.5 & 5.2  & - & 81.5 \\
		T2T-ViT-19~\cite{yuan2021tokens}  & 39.2 & 8.9   & - & 81.9 \\
		T2T-ViT-24~\cite{yuan2021tokens}   & 64.1 & 14.1   & - & 82.3 \\  
		\hline
		PVT-Small~\cite{wang2021pyramid} & 24.5 & 3.8  & 820 & 79.8 \\
		PVT-Medium~\cite{wang2021pyramid} & 44.2 & 6.7  & 526 & 81.2 \\
		PVT-Large~\cite{wang2021pyramid} & 61.4 & 9.8  & 367 & 81.7 \\ 
		\hline
		PVTv2-B0~\cite{pvtv2} & 3.4 & 0.6  & - & 70.5 \\
		PVTv2-B2~\cite{pvtv2} & 25.4 & 4.0  & - & 82.0 \\
		PVTv2-B4~\cite{pvtv2} & 62.6 & 10.1  & - & 83.6 \\ 
		\hline
		Swin-T~\cite{liu2021swin}  & 29 & 4.5  & 755 & 81.3 \\
		Swin-S~\cite{liu2021swin}  & 50 & 8.7  &  437 & 83.0 \\
		Swin-B~\cite{liu2021swin}  & 88 & 15.4  &  278 & 83.3 \\ 
		\hline
		TNT-S~\cite{tnt} & 23.8 & 5.2  & 428 & 81.5  \\
		TNT-S-2~\cite{tnt} & 22.4 & 4.7  & 704 & 81.4  \\
		TNT-B~\cite{tnt} & 65.6 & 14.1  & 246 & 82.9  \\ 
		\hline
		\textbf{PyramidTNT-Ti}  & \textbf{10.6} & \textbf{0.6}  & \textbf{2423} & \textbf{75.2} \\
		\textbf{PyramidTNT-S}  & \textbf{32.0} & \textbf{3.3}  & \textbf{721} & \textbf{82.0} \\
		\textbf{PyramidTNT-M}  & \textbf{85.0} & \textbf{8.2}  & \textbf{413} & \textbf{83.5} \\
		\textbf{PyramidTNT-B}  & \textbf{157.0} & \textbf{16.0}  & \textbf{263} & \textbf{84.1} \\
		\Xhline{1.2pt}
	\end{tabular}
	\vspace{-1em}
\end{table}

\begin{table*}[ht]
	\caption{Object detection and instance segmentation results on COCO val2017. We compare the proposed PyramidTNT-S and PyramidTNT-M with other backbones based on RetinaNet~\cite{retinanet} and Mask R-CNN~\cite{maskrcnn} frameworks, all models are trained in ``1x" schedule. FLOPs is calculated on 1280$\times$800 input.}
	\vspace{-1em}
	\setlength\tabcolsep{4.4pt}
	\begin{tabular}{l|c|cc|ccc|c|ccc|ccc}
		\Xhline{1.2pt}
		\multirow{2}{*}{Backbone} &\multicolumn{6}{c|}{RetinaNet 1$\times$} &\multicolumn{7}{c}{Mask R-CNN 1$\times$} \\
		\cline{2-14} 
		& \# FLOPs & AP & AP$_\mathrm{50}$ & AP$_\mathrm{S}$ &AP$_\mathrm{M}$ & AP$_\mathrm{L}$ & \# FLOPs & AP$^{\rm b}$ & AP$_{50}^{\rm b}$ &AP$_{75}^{\rm b}$  &AP$^{\rm m}$ &AP$_{50}^{\rm m}$ & AP$_{75}^{\rm m}$\\
		\Xhline{1.2pt}
		
		ResNet50~\cite{resnet} & 239.3G & 36.3 & 55.3 & 19.3 & 40.0 & 48.8 & 260.1G & 38.0 & 58.6 & 41.4 & 34.4 & 55.1 & 36.7 \\
		PVT-Small~\cite{wang2021pyramid} & 226.5G & 40.4 & 61.3 & 25.0 & 42.9 & 55.7 & 245.1G & 40.4 & 62.9 & 43.8 & 37.8 & 60.1 & 40.3 \\
		CycleMLP-B2~\cite{cyclemlp} & 230.9G & 40.6 & 61.4 & 22.9 & 44.4 & 54.5 & 249.5G & 42.1 & 64.0 & 45.7 & 38.9 & 61.2 & 41.8 \\
		Swin-T~\cite{liu2021swin} & 244.8G & 41.5 & 62.1 & 25.1 & 44.9 & 55.5 & 264.0G & 42.2 & 64.6 & 46.2 & 39.1 & 61.6 & 42.0 \\
		Hire-MLP-Small~\cite{hire} & 237.6G & 41.7 & - & \textbf{25.3} & \textbf{45.4} & 54.6 & 256.2G & 42.8 & 65.0 & 46.7 & 39.3 & 62.0 & 42.1 \\
		\textbf{PyramidTNT-S} & 225.9G & \textbf{42.0} & \textbf{63.1} & 25.0 & 44.9 & \textbf{57.7} & 255.9G & \textbf{43.4} & \textbf{65.3} & \textbf{47.3} & \textbf{39.5} & \textbf{62.3} & \textbf{42.2} \\
		\Xhline{1.2pt}
	\end{tabular}
	\label{table:coco-1x}
\end{table*}

\begin{table*}[ht]
	\caption{Instance segmentation results on COCO val2017. Mask R-CNN~\cite{maskrcnn} and Cascade Mask R-CNN~\cite{cascadercnn} are trained in ``3x" schedule with multi-scale strategy.}
	\vspace{-1em}
	\setlength\tabcolsep{3.8pt}
	\begin{tabular}{l|c|ccc|ccc|c|ccc|ccc}
		\Xhline{1.2pt}
		\multirow{2}{*}{Backbone} &\multicolumn{7}{c|}{Mask R-CNN 3$\times$} &\multicolumn{7}{c}{Cascade Mask R-CNN 3$\times$} \\
		\cline{2-15} 
		& \# FLOPs & AP$^{\rm b}$ & AP$_{50}^{\rm b}$ & AP$_{75}^{\rm b}$ & AP$^{\rm m}$ & AP$_{50}^{\rm m}$ & AP$_{75}^{\rm m}$ & \# FLOPs & AP$^{\rm b}$ & AP$_{50}^{\rm b}$ &AP$_{75}^{\rm b}$  &AP$^{\rm m}$ &AP$_{50}^{\rm m}$ & AP$_{75}^{\rm m}$\\
		\Xhline{1.2pt}  
		
		ResNet50~\cite{resnet} & 260.1G & 41.0 & 61.7 & 44.9 & 37.1 & 58.4 & 40.1 & 738.7G & 46.3 & 64.3 & 50.5 & 40.1 & 61.7 & 43.4 \\
		AS-MLP-T~\cite{asmlp} & 260.1G & 46.0 & 67.5 & 50.7 & 41.5 & 64.6 & 44.5 & 739.0G & 50.1 & 68.8 & 54.3 & 43.5 & 66.3 & 46.9 \\
		Swin-T~\cite{liu2021swin} & 264.0G & 46.0 & 68.2 & 50.2 & 41.6 & 65.1 & 44.8 & 742.4G & 50.5 & 69.3 & 54.9 & 43.7 & 66.6 & 47.1 \\
		Hire-MLP-S~\cite{hire} & 256.2G & 46.2 & 68.2 & 50.9 & 42.0 & 65.6 & 45.3 & 734.6G & 50.7 & 69.4 & 55.1 & \textbf{44.2} & 66.9 & \textbf{48.1} \\
		\textbf{PyramidTNT-S} & 255.9G & \textbf{47.1} & \textbf{68.9} & \textbf{51.6} & \textbf{42.2} & \textbf{65.8} & \textbf{45.4} & 794.1G & \textbf{51.0} & \textbf{69.7} & \textbf{55.3} & \textbf{44.2} & \textbf{67.0} & \textbf{48.1} \\		
		\Xhline{1.2pt}
	\end{tabular}
	\label{table:coco-3x}
\end{table*}

\paragraph{Results.}
We show the ImageNet-1K classification results in Table~\ref{table-imagenet}. Compared to the original TNT, PyramidTNT achieves much better image classification accuracy. For instance, top-1 accuracy of PyramidTNT-S is 0.5\% higher by using 1.9B fewer FLOPs compared to TNT-S. We also compare PyramidTNT with other representative CNN, MLP and transformer based models.  From the results, we can see that PyramidTNT is the state-of-the-art vision transformer.

\subsection{Object Detection}
\paragraph{Settings.} 
The object detection and instance segmentation experiments are conducted on challenging COCO 2017 benchmark~\cite{coco}, which contains 118K training images and 5K validation images. Following PVT~\cite{wang2021pyramid} and Swin Transformer~\cite{liu2021swin}, we consider three typical object detection frameworks: RetinaNet~\cite{retinanet}, Mask R-CNN~\cite{maskrcnn} and Cascade Mask R-CNN~\cite{cascadercnn} in mmdetection~\cite{mmdetection}. Noted that the four spatial shapes of our PyramidTNT are set as $\frac{H}{8}\times \frac{W}{8}$, $\frac{H}{16}\times \frac{W}{16}$, $\frac{H}{32}\times \frac{W}{32}$, and $\frac{H}{64}\times \frac{W}{64}$, in contrast to the multi-scale feature maps produced by typical backbones. To address this discrepancy, we employ four simple upsample layers consisted of a stride-two 2$\times$2 transposed convolution, followed by batch normalization~\cite{bn} and GeLU~\cite{gelu}, and a stride-one 3$\times$3 convolution, followed by another batch normalization and GeLU. Therefore our PyramidTNT can generate feature maps with strides of 4, 8, 16, and 32 pixels, w.r.t. the input image.

In order to compare with PVT~\cite{wang2021pyramid}, CycleMLP~\cite{cyclemlp} and Hire-MLP~\cite{hire}, we conduct experiments based on RetinaNet~\cite{retinanet} and Mask R-CNN~\cite{maskrcnn}. We use AdamW optimizer with a batch size of 2 images per GPU, the initial learning rate is set to 1e-4 and divided by 10 at the 8th and the 11th epoch. The weight decay is set to 0.05. All models are trained in ``1x" schedule (\ie, 12 epochs), with single-scale strategy on 8 Tesla V100 GPUs. The input image is resized such that its shorter side has 800 pixels while its longer side does not exceed 1333 pixels during training. 

In addition, we adopt another setting following~\cite{liu2021swin,asmlp,hire}, \ie, multi-scale training strategy and ``3x" schedule, based on Mask R-CNN~\cite{maskrcnn} and Cascade Mask R-CNN~\cite{cascadercnn}. During training, the input image is resized such that its shorter side is between 480 and 800 pixels while its longer side does not exceed 1333. In the testing phase, the shorter side of the input
image is fixed to 800 pixels. We also use AdamW optimizer with batch size 16 on 8 Tesla V100 GPUs. The initial learning rate is set to 1e-4 and divided by 10 at the 27th and the 33rd epoch.

\paragraph{Results.} 
Table~\ref{table:coco-1x} reports the results of object detection and instance segmentation under ``1x" training schedule. PyramidTNT-S significantly outperforms other backbones on both one-stage and two-stage detectors with similar computational cost. For example, PyramidTNT-S based RetinaNet archive 42.0 AP and 57.7 AP$_\mathrm{L}$, surpassing the models with Swin Transformer~\cite{liu2021swin} by 0.5 AP and 2.2 AP$_\mathrm{L}$, respectively. These results indicate that the pyramid architecture of TNT can help capture better global information for large objects. We conjecture that the simple upsample strategy and smaller spatial shape of PyramidTNT withhold the AP$_\mathrm{S}$ from a large promotion.

We also report the detection results under multi-scale strategy and ``3x" training schedule in Table~\ref{table:coco-3x}. PyramidTNT-S can obtain much better AP$^{\rm b}$ and AP$^{\rm m}$ than all other counterparts on Mask R-CNN~\cite{maskrcnn} and Cascade Mask R-CNN~\cite{cascadercnn}, showing its better feature representation ability. For example, PyramidTNT-S surpasses Hire-MLP-S~\cite{hire} by 0.9 AP$^{\rm b}$ on Mask R-CNN with less FLOPs constrain.

{\small
\bibliographystyle{ieee_fullname}
\bibliography{ref}
}

\end{document}